\documentclass[10pt,twocolumn,letterpaper]{article}

\usepackage[pagenumbers]{cvpr} 

\usepackage{graphicx}
\usepackage{amsmath}
\usepackage{amssymb}
\usepackage{booktabs}
\usepackage{bbm}
\usepackage{bbding}
\usepackage[dvipsnames,svgnames,x11names]{xcolor}
\definecolor{citecolor}{HTML}{0071BC}
\usepackage[bookmarks=False,pagebackref,breaklinks,colorlinks,citecolor=citecolor]{hyperref}
\usepackage[capitalize]{cleveref}
\crefname{section}{Sec.}{Secs.}
\Crefname{section}{Section}{Sections}
\Crefname{table}{Table}{Tables}
\crefname{table}{Tab.}{Tabs.}

\def\cvprPaperID{****} 
\def\confName{CVPR}
\def\confYear{2023}

\begin{document}
\title{Feature Alignment and Uniformity for Test Time Adaptation}
\author{Shuai Wang$^{1}$ \qquad Daoan Zhang$^{2}$ \qquad Zipei Yan$^{3}$ \qquad Jianguo Zhang$^{2}$ \qquad Rui Li$^{1}$ \\
  $^1$Tsinghua University \qquad $^2$Southern University of Science and Technology \\
  $^3$Hongkong Polytechnic University \\
  {\tt\small s-wang20@mails.tsinghua.edu.cn, dawn\_change@126.com, zipei.yan@connect.polyu.hk} \\
  {\tt\small zhangjg@sustech.edu.cn, leerui@tsinghua.edu.cn}
  }   
\maketitle
\begin{abstract}
    Test time adaptation (TTA) aims to adapt deep neural networks when receiving out of distribution test domain samples. In this setting, the model can only access online unlabeled test samples and pre-trained models on the training domains. We first address TTA as a \textbf{feature revision} problem due to the domain gap between source domains and target domains. After that, we follow the two measurements \textbf{alignment} and \textbf{uniformity} to discuss the test time feature revision. For test time feature uniformity, we propose a \textbf{test time self-distillation} strategy to guarantee the consistency of uniformity between representations of the current batch and all the previous batches. For test time feature alignment, we propose a \textbf{memorized spatial local clustering} strategy to align the representations among the neighborhood samples for the upcoming batch. To deal with the common noisy label problem, we propound the entropy and consistency filters to select and drop the possible noisy labels. To prove the scalability and efficacy of our method, we conduct experiments on four domain generalization benchmarks and four medical image segmentation tasks with various backbones. Experiment results show that our method not only improves baseline stably but also outperforms existing state-of-the-art test time adaptation methods. Code is available at \href{https://github.com/SakurajimaMaiii/TSD}{https://github.com/SakurajimaMaiii/TSD}.
\end{abstract}

\section{Introduction}
\label{sec:intro}
Deep learning has achieved great success in computer vision tasks when training and test data are sampled from the same distribution \cite{he2016deep,alexnet,fcn}. However, in real-world applications, performance degradation usually occurs when training (source) data and test (target) data are collected from different distributions, \ie \textit{domain shift}. In practice, test samples may encounter different types of variations or corruptions. Deep learning models will be sensitive to these variations or corruptions, which can cause performance degradation.

To tackle this challenging but practical problem, various works have been proposed to adapt the model at \textit{test} time \cite{sun2020test,DBLP:conf/nips/LiuKDBMA21,wang2021tent,IwasawaM21,pmlr-v162-niu22a,BoudiafMAB22,Wang2022,0001WDE22,memo}. Test time training (TTT) \cite{sun2020test,DBLP:conf/nips/LiuKDBMA21} adapts model using self-supervised tasks, such as rotation classification \cite{gidaris2018unsupervised} during both training and test phases. This paradigm relies on additional model modifications in both training and test phases, which is not feasible and scalable in the real world.

Similar to TTT, test time adaptation \cite{wang2021tent} (TTA) also adapts the model \ie, updates the parameters in the test phase. But TTA does not require any specific modifications in training and requires only the pre-trained source model and unlabeled target data during the test phase, which is more practical and generalizable. In TTA, the model could be adapted with only online unlabeled data. Hence, the model trained on source data is incompatible with the target data due to the possible domain gap between source data and target data.

To deal with the above-mentioned problem, we address TTA as a \textit{representation revision} problem in this paper. In the test phase of TTA, the accessed model has already learned the feature representations specialized to source domains and may generate inaccurate representations for the target domain due to the large domain gap. It is necessary to rectify the feature representations for the target domain. To achieve better representations for the target domain, we utilize the commonly used measurements for representation quality that can be summarized to feature \textit{alignment} and \textit{uniformity} \cite{wang2020understanding,zhang2022rethinking}. Alignment refers that the similar images should have the similar representations, while uniformity means that images of different classes should be distributed as uniform as possible in the latent space. Hence, we propose to address the TTA problem from the above-mentioned properties.

Most of the previous works about TTA can be inducted from the proposed representation revision perspective. Some methods adapt the source model by conducting a feature alignment process, such as feature matching \cite{DBLP:conf/ijcai/KojimaMI22,DBLP:conf/nips/LiuKDBMA21} and predictions adjustment \cite{BoudiafMAB22}. One of the representative methods is LAME \cite{BoudiafMAB22}, which encourages neighborhood samples in the feature space to have similar predictions using Laplacian adjusted maximum-likelihood estimation. Moreover, other methods aim to make target feature more uniform in feature space, including entropy minimization \cite{wang2021tent,pmlr-v162-niu22a,memo}, prototype adjustment \cite{IwasawaM21}, information maximization \cite{liang2020we} and batch normalization statistics alignment \cite{SchneiderRE0BB20,Li2018,nado2020evaluating}. One representative method is T3A \cite{IwasawaM21}, which adjusts prototypes (class-wise centroids) to have a more uniform representation by building a support set. However, none of the method address the TTA problem from the representation alignment and uniformity simultaneously. In this paper, we identify this limitation and propose a novel method that rectifies feature representation from both two properties. We formulate the two properties in TTA as \textit{test time feature uniformity} and \textit{test time feature alignment}.

\textbf{Test Time Feature Uniformity.}
Following the feature uniformity perspective, we hope that representations of test images from different classes should be distributed as uniform as possible. However, only limited test samples can be accessed in an \textit{online} manner in the TTA setting.

To better deal with all of the samples in the target domain, we propose to introduce historical temporal information for every arriving test sample. A memory bank is built to store feature representations and logits for all arriving samples to maintain the useful information from the previous data. We then calculate the pseudo-prototypes for every class by using the logits and features in the memory bank. After that, to guarantee the uniformity for the current batch of samples, the prediction distribution of prototype-based classification and model prediction (outputs of linear classifier) should be similar, \ie, the feature distribution of the current images of one class should be consistent with the feature distribution of all the previous images of the same class. This can reduce the bias of misclassified outlier samples to form a more uniform latent space.

Motivated by this, we minimize the distance between the outputs of linear and prototype-based classifiers. This pattern is similar to self-distillation \cite{be_your_own_teacher,DBLP:conf/iccv/KimJYH21} that transfers knowledge between different layers of the same network architecture. However, unlike typical self-distillation, our method does not require any ground truth supervision. We refer to this method as \textit{Test Time Self-Distillation} (TSD).

\textbf{Test Time Feature Alignment.}
Feature alignment encourages the images from the same class to have similar feature representations in the latent space. As for TTA, pseudo labels generated by the source model may be noisy due to the domain gap. Thus, instead of aligning all the positive pairs, we propose a \textit{$K$-nearest feature alignment} to encourage features from the same class to be closed or features from different classes to be far away from each other. This can reduce the negative impact imposed by the noisy labels and maintain the alignment of images with the same semantics. Specifically, we retrieve $K$-nearest features in the memory bank for the upcoming images and add consistency regularization between the representations and logits of the images. We refer to this as \textit{Memorized Spatial Local Clustering} (MSLC). The ablation of hyperparameter $K$ is shown in Table \ref{tab:ablation_study} and Fig. \ref{fig:hyper}.

\textbf{Entropy Filter and Consistency Filter.}
During the adaption, we use stored pseudo features and logits to compute the pseudo-prototypes. However, the noisy label problem cannot be completely alleviated despite the proposed efforts. To further reduce the impact, we adopt both \textit{entropy} and \textit{consistency} filters to filter noisy labels to boost performance. As for the entropy filter, we filter noisy features with high entropy when we compute prototypes because unreliable samples usually produce high entropy. 

In addition, the predictions of prototype-based and linear classifiers of the network for reliable samples should be consistent ideally. We use this property to filter unreliable samples and back-propagate the gradient using only reliable samples. We refer to this filter as the consistency filter. The ablation study on the two proposed filters is presented in Table \ref{tab:ablation_study}.

Finally, we demonstrate the effectiveness of our proposed approach on commonly used domain generalization benchmarks, including PACS \cite{li2017deeper}, VLCS \cite{torralba2011unbiased}, OfficeHome \cite{venkateswara2017deep} and DomainNet \cite{PengBXHSW19}. Furthermore, to prove the efficacy of our method, we conduct more experiments on four cross-domain medical image segmentation benchmarks, including prostate segmentation \cite{litjens2014evaluation,Antonelli2022}, cardiac structure segmentation \cite{Bernard2018,Zhuang2019,Zhuang2016} and optic cup/disc segmentation \cite{Orlando2020,Fumero2011,sivaswamy2015comprehensive}. Our method achieves the state-of-the-art performance on the above benchmarks.

We summarize our contributions as follows:
\begin{itemize}
    \item We propose a new perspective for test time adaptation from the view of feature alignment and uniformity. The proposed test time feature uniformity encourages the representations of the current batch of samples along with the uniformity of all the previous samples. The test time feature alignment manipulates the representation of the test sample according to its neighbors in the latent space to align the representations based on the pseudo label.  
    \item Specifically, to meet the online setting and noisy label problem in TTA,  we propose two complementary strategies: unsupervised self-distillation for test time feature uniformity and memorized spatial local clustering for test time feature alignment. We also propound the entropy filter and consistency filter to further mitigate the effect of the noisy labels.
    \item The experiments demonstrate that our proposed method outperforms existing test time adaptation approaches on both the domain generalization benchmarks and medical image segmentation benchmarks.
\end{itemize}

\section{Related Work}
\subsection{Domain Generalization}
Domain generalization (DG) aims to learn knowledge from multi-source domains and generalize to the unseen target domain. A primary strategy of DG is domain-invariant feature learning which aims to reduce domain gaps, including aligning distributions among multiple domains with contrastive learning \cite{DBLP:conf/iccv/MotiianPAD17,DBLP:conf/iccv/KimYPKL21,DBLP:conf/cvpr/YaoBZZSCL022,zhang2023aggregation}, learning useful representations with self supervise learning \cite{DBLP:conf/cvpr/CarlucciDBCT19}, matching statistics of feature distributions across domains \cite{tzeng2014deep,DBLP:conf/eccv/SunS16}, domain adversarial learning \cite{DANN,DBLP:conf/cvpr/LiPWK18} and causality inference \cite{DBLP:conf/cvpr/LvLLZLWL22,DBLP:conf/icml/MahajanTS21}. Meta-learning based methods \cite{DBLP:conf/nips/BalajiSC18,DBLP:conf/nips/DouCKG19,DBLP:conf/aaai/LiYSH18} simulate domain shift by dividing meta-train and meta-test domains from the original source domains. Data augmentation-based methods effectively solve the DG problem by diversifying training data \cite{HendrycksMCZGL20,YunHCOYC19,mixup,VermaLBNMLB19,mixstyle,NurielBW21,LiDGLSD22}.

Unlike the DG paradigm that focuses on training a generalizable model during the training phase with labeled source data, test time adaptation aims to adapt a pre-trained model in the test phase by using the online unlabeled target data.
\subsection{Test Time Adaptation}
Our work is mostly related to test time adaptation \cite{wang2021tent,IwasawaM21,pmlr-v162-niu22a,BoudiafMAB22,0001WDE22} or test time training \cite{sun2020test,DBLP:conf/nips/LiuKDBMA21}. Test Time Training (TTT) \cite{sun2020test,DBLP:conf/nips/LiuKDBMA21} adapts the model during the test phase via the self-supervise task, such as rotation classification \cite{gidaris2018unsupervised}. However, TTT needs to optimize the source model by jointly training supervised loss and self-supervised loss, which may not be feasible in the real world. Different from this, test time adaptation \cite{wang2021tent} aims to adapt the model during the test phase without changing the training process. Tent \cite{wang2021tent}  minimizes entropy to update the trainable parameters in Batch Normalization \cite{IoffeS15} layer. SHOT \cite{liang2020we} combines information maximization and pseudo labeling. There are some works \cite{pmlr-v162-niu22a,Wang2022} combing test time adaptation and continual learning to maintain the performance on the source domain. LAME \cite{BoudiafMAB22} adapts the output of the model rather than parameters with a laplacian adjusted maximum-likelihood estimation. There are also some works using test time Batch Normalization statistics \cite{SchneiderRE0BB20,Li2018,nado2020evaluating}, self-training \cite{0001WDE22} and feature alignment \cite{DBLP:conf/ijcai/KojimaMI22,DBLP:conf/nips/LiuKDBMA21}. The above methods could be viewed as either feature uniformity or feature alignment as discussed in the Sec. \ref{sec:intro}, while our method benefits from both properties.

\begin{figure*}
    \setlength{\belowcaptionskip}{-10pt}
    \centering
    \includegraphics[width=0.75\linewidth]{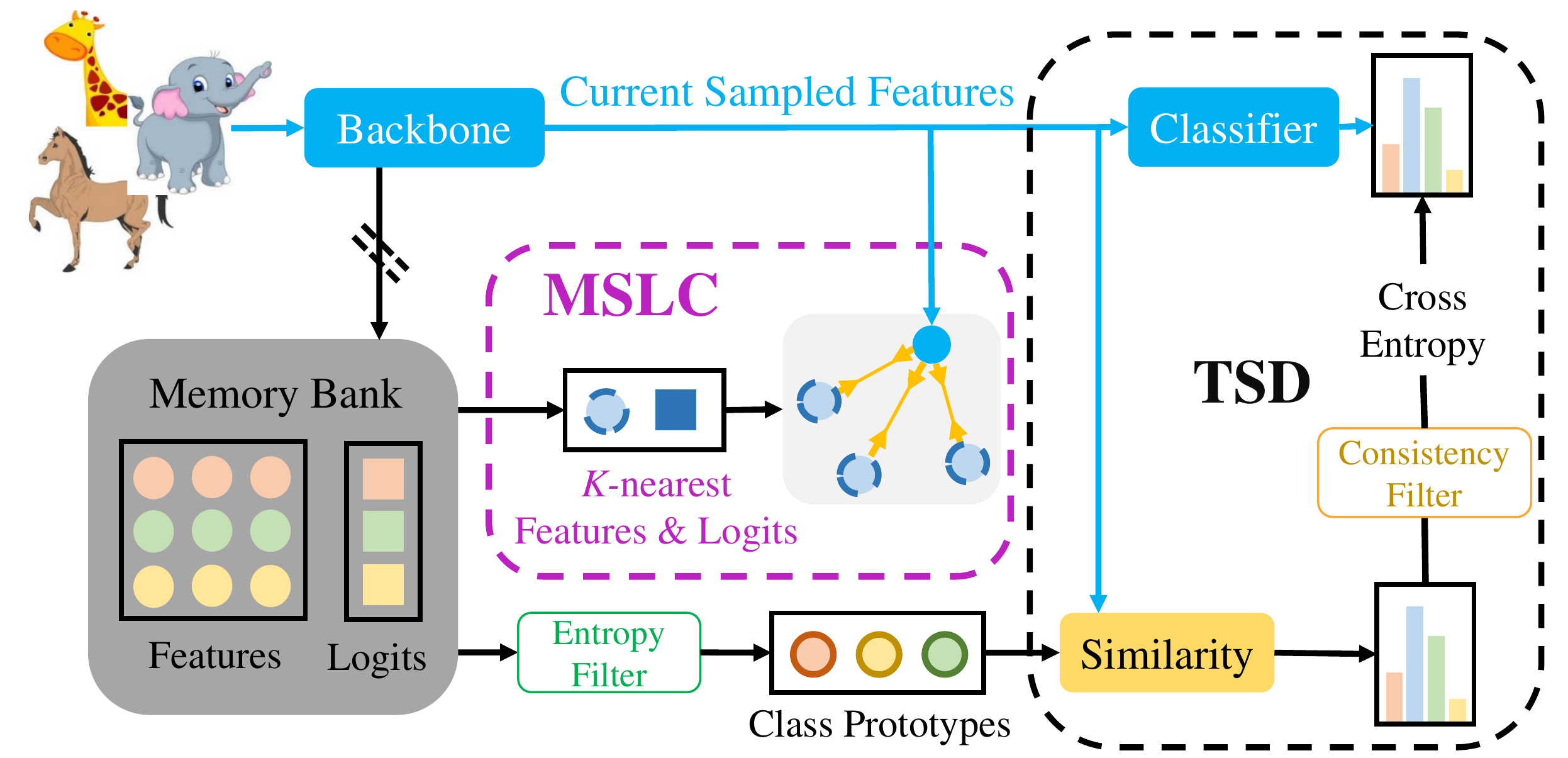}
    \caption{\textbf{Overview of our proposed method}. Blue lines denote forward and backward and black lines denote only forward (\ie without gradient backpropagation). Different colours of features, logits and prototypes mean different classes. MSLC: Memorized Spatial Local Clustering. TSD: Test Time Self-Distillation.}
    \label{fig:framework}
    \vspace{-0.5em}
\end{figure*}
\section{Method}
Fig. \ref{fig:framework} illustrates the overall pipeline of our method. We describe the details of the problem settings and our method in this section. 
\subsection{Preliminaries}
In test time adaptation (TTA), we can only achieve target domain unlabeled images in an online manner and pre-trained model on the source domain. The source model is trained with standard empirical risk minimization on source domains, \eg cross entropy loss for the image classification task. Given the model trained on $\mathcal{D}_s$, we aim to adapt the model using unlabeled target data $\{ x_{i} \} \in \mathcal{D}_t$, $i \in \{1...N\}$, where $x_i$ denotes the $i$th image of target domain $\mathcal{D}_t$ and $N$ denotes the number of target images, $\mathcal{D}_s$ denotes the source domain. During testing, we initialize the model $g=f\circ h$ with source model parameters trained on source domain $\mathcal{D}_s$, where $f$ denotes backbone and $h$ denotes linear classification head. The output of model $g$ for image $x_i$ is denoted as $p_i = g(x_i) \in \mathbb{R}^C$ where $C$ is the number of class.
\subsection{Test Time Self-Distillation}
\label{sec:selfdist_bank}
During adaption, given a batch of unlabeled test samples, we can generate image embeddings $z_i=f(x_i)$, logits $p_i=h(z_i)$ and pseudo labels $\hat{y}_i = \mathop{\arg\max} p_i$ via the pre-trained model.

We then maintain a memory bank $\mathcal{B}=\{(z_i,p_i)\}$ to store image embeddings $z_i$ and logits $p_i$. Following T3A \cite{IwasawaM21}, the memory bank is initialized with the weights of the linear classifier. When the target sample $x_i$ comes, for every image, we add the image embedding $z_i$ and logits $p_i$ into the memory bank. To build up the relations between the current samples and all of the previous samples, the pseudo-prototypes shall be generated for each class. 
The prototype of class $k$ could be formulated as
\begin{equation}
    c_k = \frac{\sum_{i}z_i\mathbbm{1}[\hat{y}_i=k]}{\sum_{i}\mathbbm{1}[\hat{y}_i=k]},
    \label{eq:proto}
\end{equation}
where $\mathbbm{1}(\cdot)$ is an indicator function, outputting value 1 if the argument is true or 0 otherwise.
However, some pseudo labels may be assigned to the wrong class, leading to incorrect prototype computation. We use Shannon entropy \cite{Shannon48} filter to filter noisy labels. For prediction $p_i$, its entropy could be computed as $H\left( p_i \right) = -\sum \sigma (p_i)\log \sigma (p_i)$ where $\sigma$ denotes the softmax operation. We aim to filter unreliable features or predictions with high entropy because lower entropy usually means higher accuracy \cite{wang2021tent}.
Specifically, for every class, image embeddings with the top-$M$ highest entropy in the memory bank would be ignored. After that, we use filtered embeddings to calculate prototypes as Eq. \ref{eq:proto} and define prototype-based classification output as the softmax over the feature similarity to prototypes for class $k$:
\begin{equation}
    y_i^k = \frac{\exp{(\text{sim}(z_i,c_k))}}{\sum_{k^{\prime}=1}^{C}\exp{(\text{sim}(z_i,c_{k^{\prime}}))}},
\end{equation}
where $\text{sim}(z_i,c_k)$ denotes cosine similarity between $z_i$ and $c_k$. 

The prototype-based classification results $y_i$ and outputs $p_i$ of the network $g$ should share a similar distribution for the same input. Thus, the loss to maintain uniformity is proposed as
 \begin{equation}
    \mathcal{L}_i(p_i,y_i) = -\sigma (p_i)\log y_i.
    \label{eq:self_distillation}
\end{equation}
Note that $p_i$ is a soft pseudo label rather than a hard pseudo label. The reason for using a soft label is that a soft label usually provides more information \cite{hinton2015_kd}. By using the proposed test time self-distillation, the network could map the uniformity for the current samples to improve the quality of representations.

Although we use an entropy filter to drop noisy labels when computing prototypes, there are still some mistake predictions inevitable. We propose that, for a reliable sample, the outputs of the linear fully connected layer and prototype-based classifier should be similar. Hence, we adopt the consistency filter to identify the mistake predictions. Specially, if the linear classifier and prototype-based classifier produce the same predictions, \ie the same result after executing $argmax$ to the logits, we assume that this sample is reliable. This strategy could be implemented using a filter mask for image $x_i$ as follows
\begin{equation}
    \mathcal{M}_i = \mathbbm{1}[\mathop{\arg\max} p_i = \mathop{\arg\max} y_i].
\end{equation}
By conducting a consistency filter, we further filter unreliable samples and the unsupervised self-distillation loss could be formulated as follows
\begin{equation}
    \mathcal{L}_{tsd} = \frac{\sum_{i}\mathcal{L}_i*\mathcal{M}_i}{\sum_{i}\mathcal{M}_i}.
    \label{eq:self}
\end{equation}

\subsection{Memorized Spatial Local Clustering}
\label{sec:mslc}
As mentioned before, features belonging to the same class should be aligned in the latent space. However, this situation may differ in TTA due to the domain gap between the target and source domains. We encourage $K$-nearest neighborhood features instead of all the features to be close to reducing the noisy label impact. A simple strategy is to add consistency regularization within a batch of samples. However, historical temporal information is ignored and the alignment is less effective. Moreover, there is a trivial solution that the model can easily map all images to a certain class if we use only one batch sample for alignment \cite{Belkin2003}.

To handle these problems, we concatenate spatial local clustering with the memory bank. We begin from retrieving $K$-nearest features in the memory bank for image $x$. Based on our assumption, the logits of image $x$ should be aligned with the logits of its nearest neighbors in the latent space. To achieve this, we align the two varieties of logits according to the distance between the image embeddings of image $x$ and its neighbors. The formulation is presented as follows
\begin{equation}
    \vspace{-0.5em}
    \mathcal{L}_{mslc} = \frac1K\sum^{K}_{j=1}\text{sim}(z,z_j) (\sigma (p)-\sigma (p_j))^2,
    \label{eq:spatial_loss}   
    \vspace{-0.3em}
\end{equation}
where $\text{sim}(z,z_j)$ denotes cosine similarity between $z$ and $z_j$. $\{z_j\}_{j=1}^K$ denotes the $K$-nearest image embeddings of $z$ in the memory bank $\mathcal{B}$ and $p_j$ denotes the corresponding logits. If $z_j$ and $z$ are close in feature space, \ie $\text{sim}(z,z_j)$ is large, this objective function will push $p_j$ and $p$ to get close. We detach the gradient of $\text{sim}(z,z_j)$, \ie $\text{sim}(z,z_j)$ would be viewed as constant, to avoid the trivial solution that the model will output a constant result regardless of different samples.
\subsection{Training Objective Function}
Combing Eq.~\ref{eq:self} and Eq.~\ref{eq:spatial_loss}, we formulate the final objective function as
\begin{equation}
    \mathcal{L} = \mathcal{L}_{tsd} + \lambda\mathcal{L}_{mslc},
    \label{eq:loss}
\end{equation}
where $\lambda$ is the trade-off parameter to balance different loss functions. In our implementation, we use cosine similarity as the similarity metric. Specifically, we define $\text{sim}(x,y)=x^\top y/\Vert x \Vert \Vert y \Vert$.

During testing phase, the adaptation is performed in an \textit{online} manner. Specifically, when receiving image $x_T$ at time point $T$, the model state is initialized with the parameters update from the last image $x_{T-1}$. The model produces the prediction $p_T=g(x_T)$ after receiving new samples $x_T$ and updates the model using Eq. \ref{eq:loss} with only \textit{one} step gradient descent. It is noticed that the adaption could continue as long as there exists test data.
\section{Experiments}
\subsection{Experimental Setup}
\label{sec:exp_setup}
\textbf{Datasets}. \textbf{PACS} \cite{li2017deeper} contains 9,991 examples and 7 classes that are collected from 4 domains: art, cartoons, photos, and sketches. \textbf{OfficeHome} \cite{venkateswara2017deep} is consisted of 4 domains: art, clipart, product, and real, which includes 15,588 images and 65 classes. \textbf{VLCS} \cite{torralba2011unbiased} comprises four domains: Caltech101, LabelMe, SUN09 and VOC2007 and includes 10,729 images and 5 classes. \textbf{DomainNet} \cite{PengBXHSW19}, a large-scale dataset has six domains $d\in$\{clipart, infograph, painting, quickdraw, real, sketch\} with 586,575 images and 345 classes.

\textbf{Models}. In the main experiments, we evaluated different methods on ResNet-18/50 \cite{he2016deep} equipped with Batch Normalization \cite{IoffeS15}, which is widely used in domain adaptation and generalization literature. Furthermore, we tested our algorithm on different backbones, including Vision Transformer (ViT-B/16) \cite{vit}, ResNeXt-50 (32$\times$4d) \cite{resnext}, EfficientNet (B4) \cite{efficientnet} and MLP-Mixer (Mixer-L16) \cite{mlp-mixer}.

\textbf{Implementation}. For source training, we choose one domain as the target domain and the other domains as the source domains. We split all images from the source domains to 80\%/20\% for training and validation. We use the Adam optimizer \cite{Adam} with $5e^{-5}$ as the learning rate. All models are initialized with ImageNet-1K \cite{Russakovsky2015} pre-trained weights except for ViT-B/16 and MLP-Mixer when we use ImageNet-21K pre-trained weights. We use \texttt{torchvision} implementations of all models except for ViT-B/16 and MLP-Mixer; instead, we use the implementations in the \texttt{timm} library.

For test time adaptation, we use the Adam optimizer \cite{Adam} and set the batch size as 128. We empirically set the trade-off parameter $\lambda=0.1$ (\cf Eq. \ref{eq:loss}). Unlike \cite{wang2021tent,pmlr-v162-niu22a}, all the trainable layers are updated and no special selection is required in our method. We use PyTorch \cite{PaszkeGMLBCKLGA19} for all implementations. We report the accuracy of the whole target domain for evaluation. For all experiments, we report the average of three repetitions with different weight initialization, random seed and data split. Please refer to our supplementary material for more details, including detailed results and error bars.

\textbf{Hyperparameter search and model selection}. We use training domain validation \cite{gulrajani2021in} for source model training. We select the model with the highest accuracy on the validation set. For hyperparameter search, we search learning rate $lr$ in $\{1e^{-3},1e^{-4},1e^{-5},1e^{-6}\}$, the hyperparameter for feature filter $M \in \{1,5,20,50,100,\text{NA}\}$ where NA denotes no entropy filter. We emphasize that all hyperparameters in the TTA setting should be selected before accessing test samples. We do hyperparameter search on \textit{training domain validation set}.

\begin{table}[t]
    \centering
    \caption{Comparison of our method and existing test time adaptation methods with ResNet18 backbone. We highlight the \textbf{best} and the \underline{second} results.}
    \resizebox{\linewidth}{!}
    {
    \begin{tabular}{lccccc}
    \toprule
         Method & PACS & OfficeHome & VLCS & DomainNet & Avg.\\
    \midrule
    ERM \cite{DBLP:books/daglib/0097035} & 82.07 & 63.12 & 72.75 & 38.95 & 64.22\\
    BN  \cite{SchneiderRE0BB20}      & 82.82 & 62.30 & 64.31& 37.80 & 61.81\\
    Tent \cite{wang2021tent}    & \underline{84.92} & 63.75 & 67.36 & 38.95 & 63.75\\
    PL \cite{lee2013pseudo} & 84.64 & 60.22& 68.93 & 35.23 & 62.26\\ 
    SHOT-IM \cite{liang2020we}  & 82.55 & 63.42 & 64.90 & 39.50 & 62.59\\
    T3A \cite{IwasawaM21}    & 83.50 & \underline{64.25} & \underline{73.03} & \underline{39.61} & \underline{65.10} \\
    ETA \cite{pmlr-v162-niu22a}    & 82.70 & 62.46& 64.35& 39.43 & 62.24\\
    LAME \cite{BoudiafMAB22}    & 84.58 & 62.20& 72.88 & 37.49 & 64.29\\
    \hline
    Ours     & \textbf{87.32} & \textbf{64.83} & \textbf{73.61} & \textbf{40.19} & \textbf{66.49} \\
    \bottomrule
    \end{tabular}
    }
    \label{tab:results_resnet18}
\end{table}
\begin{table}[t]
    \centering
    \caption{Comparison of our method and existing test time adaptation methods with ResNet50 backbone.  We highlight the \textbf{best} and the \underline{second} results.}
    \resizebox{\linewidth}{!}{
    \begin{tabular}{lccccc}
    \toprule
        Method & PACS & OfficeHome & VLCS & DomainNet & Avg. \\
    \midrule
    ERM \cite{DBLP:books/daglib/0097035}& 84.59 & 67.37 & \underline{74.01} & 45.20 & 67.74\\
    BN \cite{SchneiderRE0BB20}      & 85.03 & 66.10 & 64.78 & 43.38 & 64.82\\
    Tent \cite{wang2021tent}     & \underline{87.48} & 67.96 & 69.20 & 44.71 & 67.34\\
    PL \cite{lee2013pseudo}       & 85.23 & 67.13 & 68.52 & 41.18 & 65.52\\ 
    SHOT-IM \cite{liang2020we}  & 85.50 & 67.39 & 65.23 & \underline{46.30} & 66.11 \\
    T3A \cite{IwasawaM21} & 86.04 & \underline{68.29} & 73.98 & 46.16 & \underline{68.62}\\
    ETA \cite{pmlr-v162-niu22a}      & 85.04 & 66.21& 64.79 & 46.13 & 65.54\\
    LAME \cite{BoudiafMAB22}     & 86.62 & 66.19& 73.94  & 43.20 & 67.49\\
    \hline
    Ours     & \textbf{89.41} & \textbf{68.67} & \textbf{74.52} & \textbf{47.73} & \textbf{70.08}\\
    \bottomrule
    \end{tabular}}
    \label{tab:results_resnet50}
\end{table}

\textbf{Baselines}. We compared our method with Empirical Risk Minimization (ERM) \cite{DBLP:books/daglib/0097035}, Tent \cite{wang2021tent}, T3A \cite{IwasawaM21}, SHOT-IM \cite{liang2020we}, ETA \cite{pmlr-v162-niu22a}, Test Time Batch Normalization (BN) \cite{SchneiderRE0BB20}, Laplacian Adjusted Maximumlikelihood Estimation (LAME) \cite{BoudiafMAB22} and PseudoLabeling (PL) \cite{lee2013pseudo}. We use the implementation from the released code of \texttt{T3A} library\footnote{\href{https://github.com/matsuolab/T3A}{https://github.com/matsuolab/T3A}}, except for LAME and ETA, we use the source code of authors.
\subsection{Comparative Study}

\textbf{Comparison with TTA methods}.  Table \ref{tab:results_resnet18} and Table \ref{tab:results_resnet50} present the results on ResNet-18/50 of four different datasets. From Table \ref{tab:results_resnet18} and \ref{tab:results_resnet50}, we can see that our method generally achieves the state-of-the-art performance. Specifically, the proposed method improves the ERM \cite{DBLP:books/daglib/0097035} baseline with 4.8\%, 1.3\%, 0.5\%, 2.53\% for each dataset,  respectively. Other test time adaptation methods do not improve the baseline as stably as ours.  

\begin{figure}[t]
    \centering
    \includegraphics[width=0.5\textwidth]{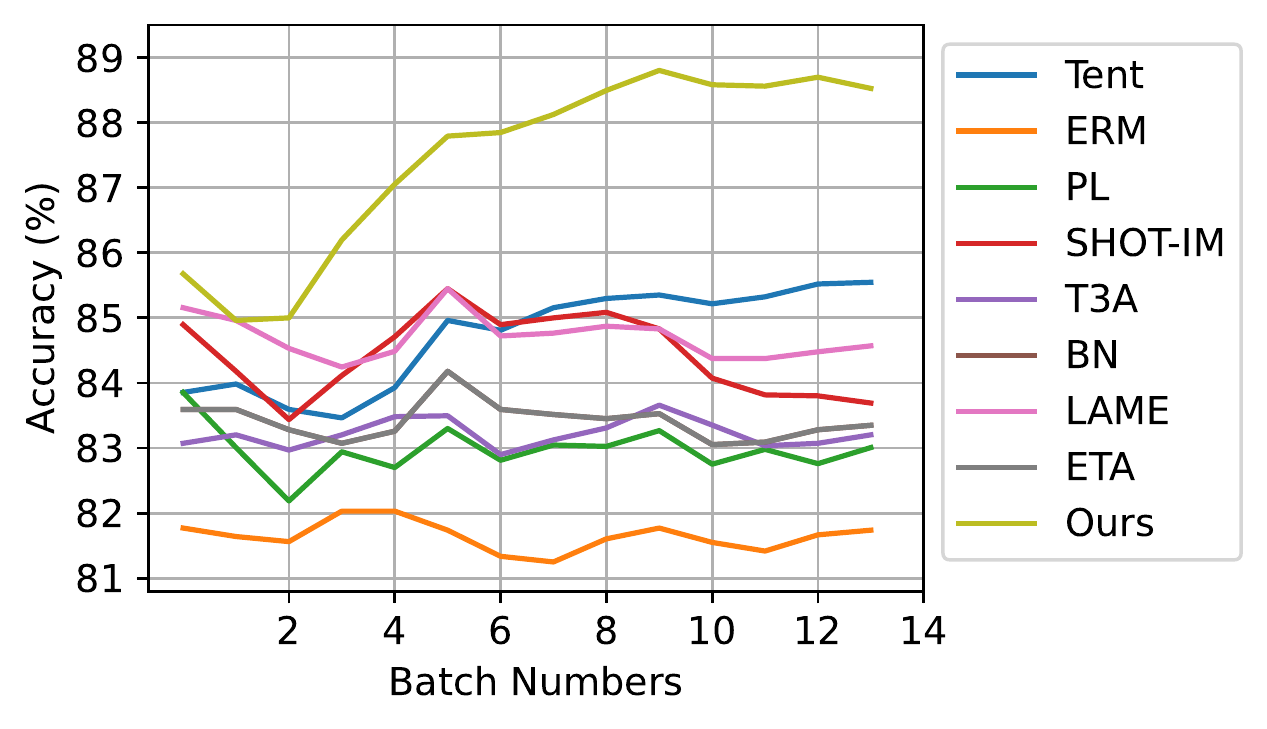}
    \caption{Accuracy visualizations of different methods during the adaptation process on the PACS dataset (target domain: art). The ``Batch Numbers'' indicates how many batches of images the model has been updated with.}
    \label{fig:process}
\end{figure}

We also visualized the change in accuracy of different methods throughout the adaption process which is shown in Fig. \ref{fig:process}. The ``Batch Numbers''  in Fig. \ref{fig:process} indicates how many batches of images that the model has been updated with. We can see that our method could adapt the data much faster and achieve higher accuracy on the target domain.

\textbf{Comparison with DG/SFDA methods}. The above experiments mainly focus on test time adaptation that aims to adapt the model during \textit{test} phase. It is natural to ask: \textit{how about our method compared with domain generalization or source free domain adaptation methods?} To answer this question, we first compared our method with some recent domain generalization or source free domain adaptation methods, such as SWAD \cite{swad}, PCL \cite{DBLP:conf/cvpr/YaoBZZSCL022}, DNA \cite{DBLP:conf/icml/ChuJZWWZM22}, and F-mix \cite{KunduKBMKJR22}\footnote{``F-mix'' denotes \textit{feature-mixup} in \cite{KunduKBMKJR22}.} on PACS and DomainNet dataset. From Table \ref{tab:compared_dg}, we can see that our method outperforms the state-of-the-art methods in domain generalization. Furthermore, combining SWAD \cite{swad}, we achieve the impressive 91\% accuracy with ResNet50 backbone.

We also report the result of the challenging DomainNet dataset. The results are listed in Table \ref{tab:compared_sfda}. It can be seen that our method outperforms the state-of-the-art methods of domain generalization, such as SWAD \cite{swad} and DNA \cite{DBLP:conf/icml/ChuJZWWZM22}. Also, our method could significantly improve SWAD \cite{swad}, which outperforms the current state-of-the-art SFDA method \cite{KunduKBMKJR22}. Note that online test time adaptation is more flexible in the real world because SFDA adapts the test data in an offline manner which requires more training loops and resources than TTA. 

To summarize, our model can outperform all the domain generalization and test time adaptation methods on various datasets.

\begin{table}[t]
  \centering
  \caption{Compare with domain generalization methods on PACS dataset with ResNet50. $\dag$: numbers are from the original literature. $\ddag$ denotes our implementation results. We highlight the \textbf{best} and the \underline{second} results in each column.}
    \begin{tabular}{lccccc}
    \toprule
        Method  & A & C & P& S & Avg. \\
    \midrule
    ERM \cite{DBLP:books/daglib/0097035}   & 82.5  & 80.8  & 94.1  & 81.0    & 84.6 \\
    $\text{DNA}^{\dag}$ \cite{DBLP:conf/icml/ChuJZWWZM22}& 89.8  & 83.4  & \underline{97.7}  & 82.6  & 88.4 \\
    $\text{PCL}^{\dag}$ \cite{DBLP:conf/cvpr/YaoBZZSCL022}  & \underline{90.2}  & 83.9  & \textbf{98.1} & 82.6  & 88.7 \\
    $\text{SWAD}^{\ddag}$ \cite{swad} & 89.3  & 83.2  & 96.9  & 83.4  & 88.2 \\
    \hline
    Ours & 87.7  & \underline{88.8}  & 96.2  & \underline{85.0}    & \underline{89.4} \\
    SWAD + Ours & \textbf{92.2} & \textbf{89.2} & 97.1  & \textbf{85.6} & \textbf{91.0} \\
    \bottomrule
    \end{tabular}%
  \label{tab:compared_dg}%
\end{table}%

\begin{table}[t]
  \centering
  \caption{Compare with domain generalization methods and source free domain adaptation methods on DomainNet dataset with ResNet50. $\dag$: numbers are from the original literature. $\ddag$ denotes our implement results. We highlight the \textbf{best} and the \underline{second} results in each column.}
   \resizebox{\linewidth}{!}{
    \begin{tabular}{lccccccc}
    \toprule
    Method      &clip & info & paint & quick & real & sketch & Avg.\\
    \midrule
    \multicolumn{8}{l}{Domain generalization methods} \\
    ERM \cite{DBLP:books/daglib/0097035}  & 64.8 & 22.1 & 51.8 & 13.8 & 64.7 & 54.0 & 45.2 \\
    $\text{PCL}^{\dag}$ \cite{DBLP:conf/cvpr/YaoBZZSCL022}   & 67.9  & 24.3  & 55.3  & 15.7  & 66.6  & 56.4  & 47.7 \\
    $\text{DNA}^{\dag}$ \cite{DBLP:conf/icml/ChuJZWWZM22}  & 66.1  & 23.0    & 54.6  & 16.7  & 65.8  & 56.8  & 47.2 \\
    $\text{SWAD}^{\ddag}$ \cite{swad} & 66.1  & 22.4  & 53.6  & 16.3  & 65.5  & 56.2  & 46.7 \\
    \midrule
    \multicolumn{8}{l}{Source free domain adaptation methods} \\
    $\text{F-mix}^{\dag}$ \cite{KunduKBMKJR22} & \textbf{75.4} & \underline{24.6}  & \underline{57.8}  & \underline{23.6}  & 65.8  & \underline{58.5}  & \underline{51.0} \\
    \midrule
    Ours & 66.1 & 24.1 & 52.8 & 18.2 & \textbf{68.5} & 56.7 & 47.7\\
    SWAD + Ours & \underline{69.2}  & \textbf{28.4} & \textbf{58.2} & \textbf{26.2} & \underline{68.1}  & \textbf{59.6} & \textbf{51.6} \\
    \bottomrule
    \end{tabular} }
  \label{tab:compared_sfda}%
\end{table}

\subsection{Ablation Study}
In this section, we mainly conduct various ablation studies on the PACS dataset with ResNet50 backbone.

\textbf{Effectiveness of key components}. Table \ref{tab:ablation_study} shows the contribution of each term mentioned in our method. Compared with the ERM baseline, self-distillation improves the baseline 3.2\%. This shows that our method could efficiently capture historical temporal features to maintain uniformity. The entropy filter and consistency filter improve the performance with 0.6\% and 0.5\%, respectively, which suggests that they mitigate the problem of noise pseudo labels. Finally, our proposed memorized spatial local clustering brings a 0.5\% gain which suggests that the module also helps with the revision of representations.

\begin{figure}[t]
    \centering
    \includegraphics[width=0.49\textwidth]{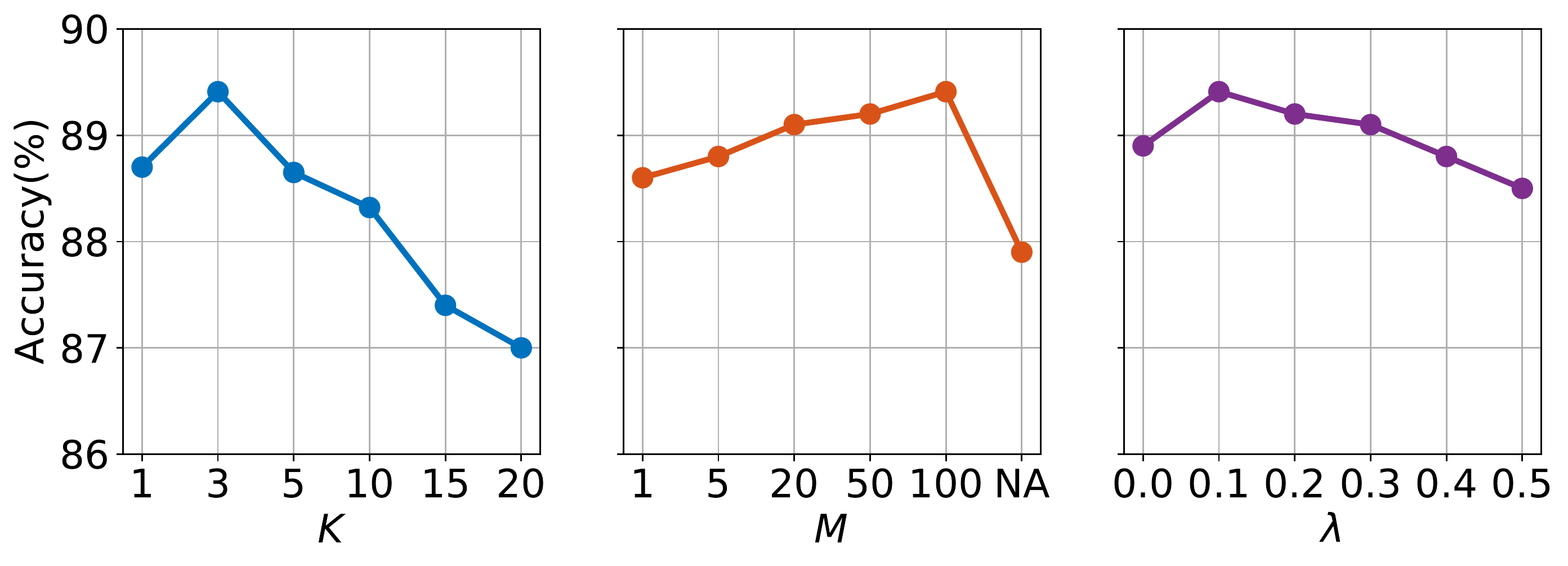}
    \caption{Sensitivity analysis about the number of nearest neighbors $K$ in memorized spatial local clustering, entropy filter hyperparameter  $M$ and trade-off parameter $\lambda$ (\cf Eq. \ref{eq:loss}).}
    \label{fig:hyper}
    \vspace{-1em}
\end{figure}

\textbf{Sensitivity to hyperparameter}. Our method has three hyperparameters: top-$M$ features for entropy filtering to compute prototypes, the number of nearest neighbors $K$ in memorized spatial local clustering and the trade-off parameter $\lambda$ in Eq. \ref{eq:loss}. Fig. \ref{fig:hyper} presents the ablation study on hyperparameters. First, for the number of nearest neighbors $K$, when $K\in\{1,3,5\}$ achieves good performance. However, though it stills improves the ERM baseline, $K\in\{10,15,20\}$ leads to performance degradation. We conjecture that using too many neighbors may increase the probability of introducing misclassified features. Second, for the entropy filter hyperparameter $M$, we found that the proposed method is robust to the choice of $M$. The performance improves stably with the $M$ increasing from 1 to 100. If we mute the entropy filter, our method still achieves a competitive performance. Third, for trade-off parameter $\lambda$, $\lambda\in[0.1,0.5]$ achieves good performance and $\lambda=0.1$ achieves the best performance.

\subsection{Analysis}
\begin{table}[t]
  \centering
  \caption{Ablation Study. SD means self-training using self-distillation (\cf Eq \ref{eq:self_distillation})  without any filter. EF and CF mean entropy filter and consistency filter. MSLC means memorized spatial local clustering loss (\cf Eq. \ref{eq:spatial_loss}).}
    \begin{tabular}{cccccl}
    \toprule
    \#    & SD & EF & CF  & MSLC  & Acc(\% $\uparrow$) \\
    \midrule
    0     &       &       &       &       & 84.59  \\
    1     & \checkmark     &       &       &       & 87.80 (\textcolor{red}{+3.21})\\
    2     & \checkmark     & \checkmark    &       &       & 88.40 (\textcolor{red}{+3.81})\\
    3     & \checkmark    & \checkmark     & \checkmark     &       & 88.92 (\textcolor{red}{+4.33})\\
    4     & \checkmark     & \checkmark     & \checkmark     & \checkmark     & 89.41 (\textcolor{red}{+4.82})\\
    \bottomrule
    \end{tabular}%
  \label{tab:ablation_study}
\end{table}

\begin{table}[t]
    \centering
    \caption{Results with different backbones. \textbf{Bold} type indicates performance improvement compared with baseline. All baseline models are trained in a standard ERM manner.}
    \resizebox{\linewidth}{!}{
    \begin{tabular}{lccc}
    \toprule
           Backbones & PACS & OfficeHome & VLCS \\
    \midrule
    ResNet18 \cite{he2016deep}     & 82.07 & 63.12 & 72.75 \\
    +Ours & \textbf{87.32} & \textbf{64.83} & \textbf{73.61}\\
    \hline
    ResNet50 \cite{he2016deep} & 84.59 & 67.37 & 74.01\\
    +Ours   & \textbf{89.41} & \textbf{68.67} & \textbf{74.52}\\
     \hline
    ResNeXt-50 \cite{resnext} & 86.67 & 72.66 & 78.50 \\
    +Ours & \textbf{91.33} & \textbf{74.18} & \textbf{79.38}\\
    \hline
    ViT-B/16 \cite{vit}     & 87.13 & 79.06 & 78.70 \\
    +Ours & \textbf{90.20} & \textbf{81.80} & \textbf{79.90}\\
    \hline
    EfficientNet-B4 \cite{efficientnet} & 85.11 & 74.65 & 77.14\\
    +Ours & \textbf{85.41} & 72.24 &  \textbf{79.42}\\
    \hline
    Mixer-L16 \cite{mlp-mixer} & 84.59  & 71.36 & 76.53\\
    +Ours & \textbf{88.47} &   \textbf{74.82} & \textbf{79.75}\\
    \bottomrule
    \end{tabular}}
    \label{tab:results_backbones}
\end{table}

\textbf{Scalability on different models}. We validated our method on various backbones to verify that our method does not rely on any specific network structure design. Table \ref{tab:results_backbones} shows that the proposed method generally improves baseline performance regardless of different backbones. For example, for ViT-B/16, the proposed method improves ERM \cite{DBLP:books/daglib/0097035} 3.1\%, 2.7\%, 1.2\% for different datasets, respectively. This show that our method is `` plug and play " for different deep neural networks which is important for applications in the real world.

\begin{figure}[t]
    \centering
    \begin{subfigure}[t]{0.22\textwidth}
            \centering
            \includegraphics[width=\textwidth]{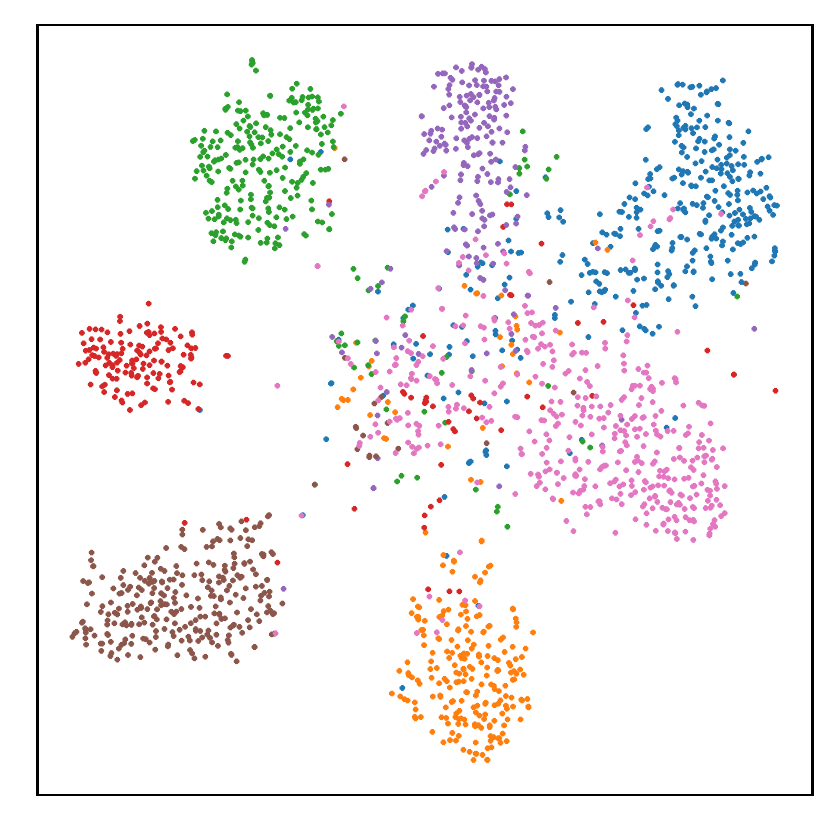}
            \caption{ERM}
            \label{fig:tsne_erm}
    \end{subfigure}
    \begin{subfigure}[t]{0.22\textwidth}
            \centering
            \includegraphics[width=\textwidth]{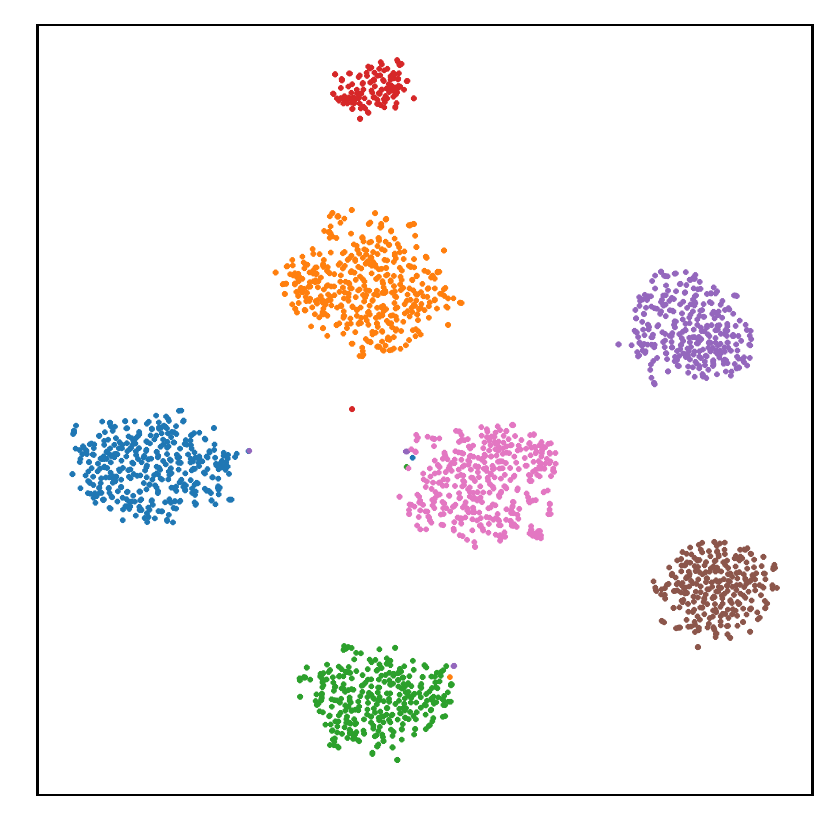}
            \caption{Our method}
            \label{fig:tsne_ours}
    \end{subfigure}
    \caption{tSNE \cite{tsne} visualization of learned feature embeddings. (a) ERM baseline without adaptation. (b) After adaptation with our method. Different colours denote different classes.}
    \vspace{-5pt}
    \label{fig:tsne}
\end{figure}

\textbf{Qualitative analysis by tSNE}. We provide tSNE \cite{tsne} visualizations of the ERM baseline and our method, as shown in Fig. \ref{fig:tsne}. The learned features of the ERM baseline on the target domain are not well separated due to the large domain gap, as shown in Fig. \ref{fig:tsne_erm}. After adaptation to the target domain, our method could generate more uniform and alignment features, as shown in Fig. \ref{fig:tsne_ours}.

\begin{figure}[t]
    \centering
    \begin{subfigure}[t]{0.16\textwidth}
            \centering
            \includegraphics[width=\textwidth]{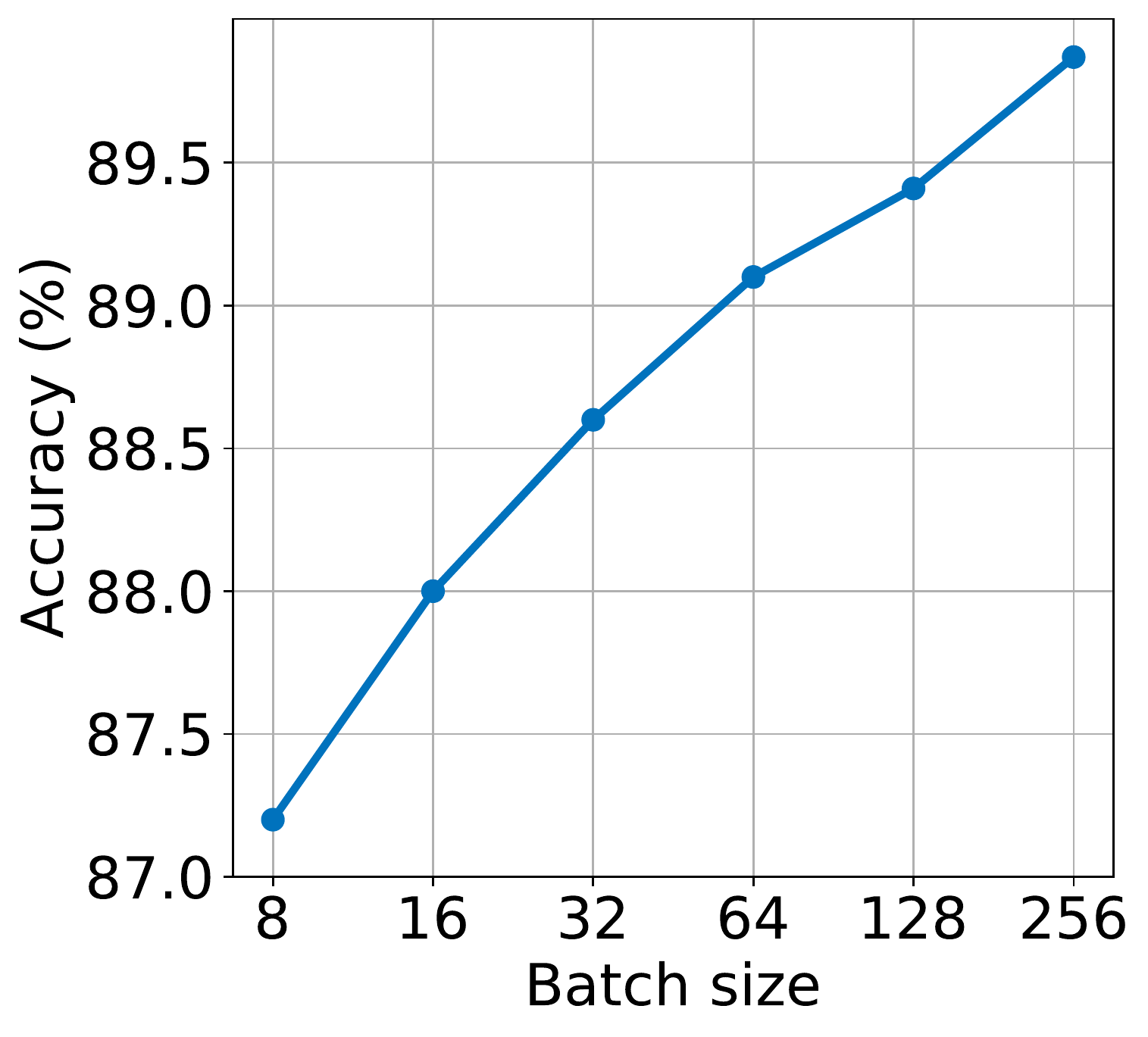}
            \caption{Accuracy}
            \label{fig:bs_acc}
    \end{subfigure}
    \begin{subfigure}[t]{0.153\textwidth}
            \centering
            \includegraphics[width=\textwidth]{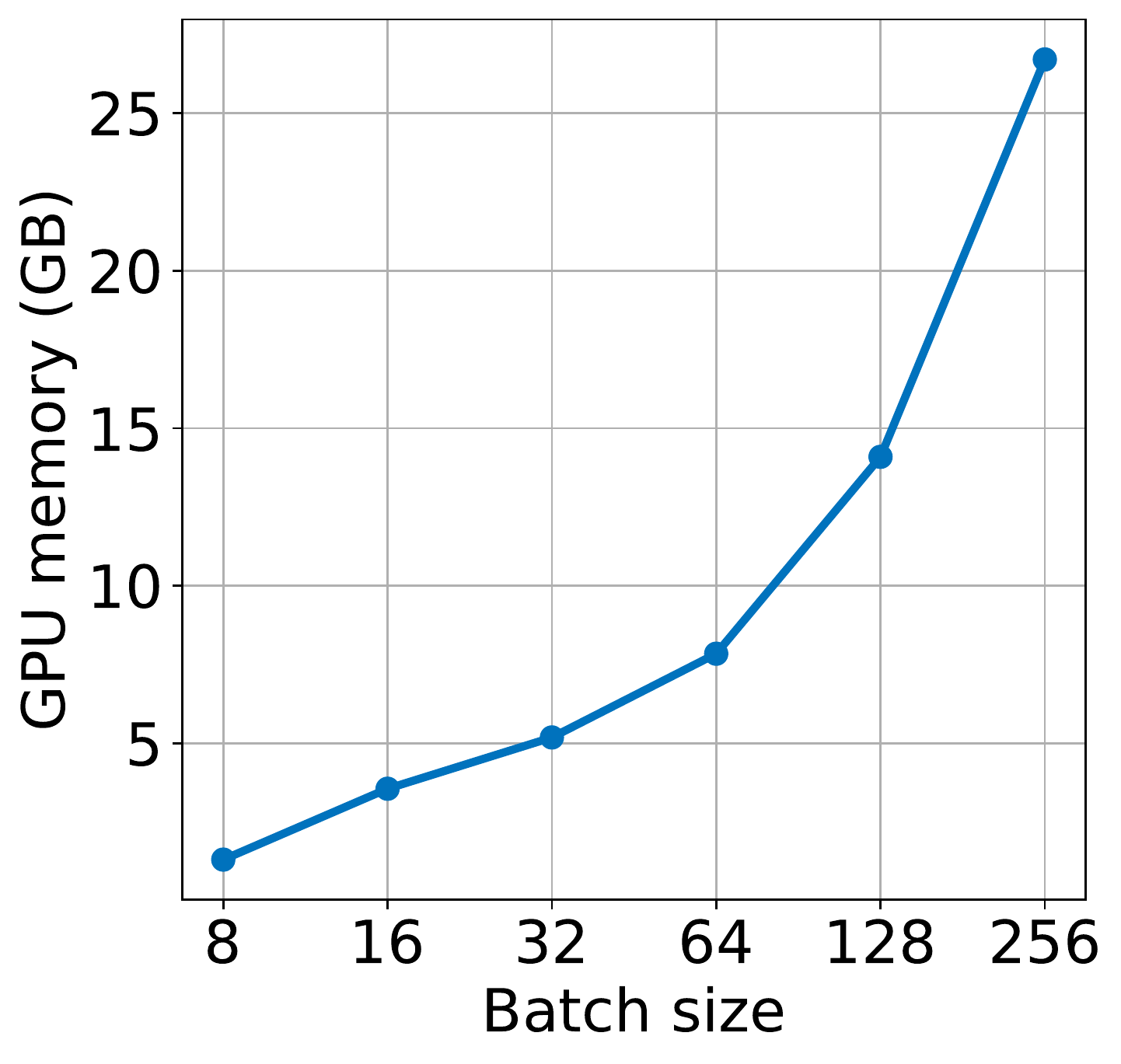}
            \caption{GPU memory}
            \label{fig:bs_memory}
    \end{subfigure}
    \begin{subfigure}[t]{0.153\textwidth}
            \centering
            \includegraphics[width=\textwidth]{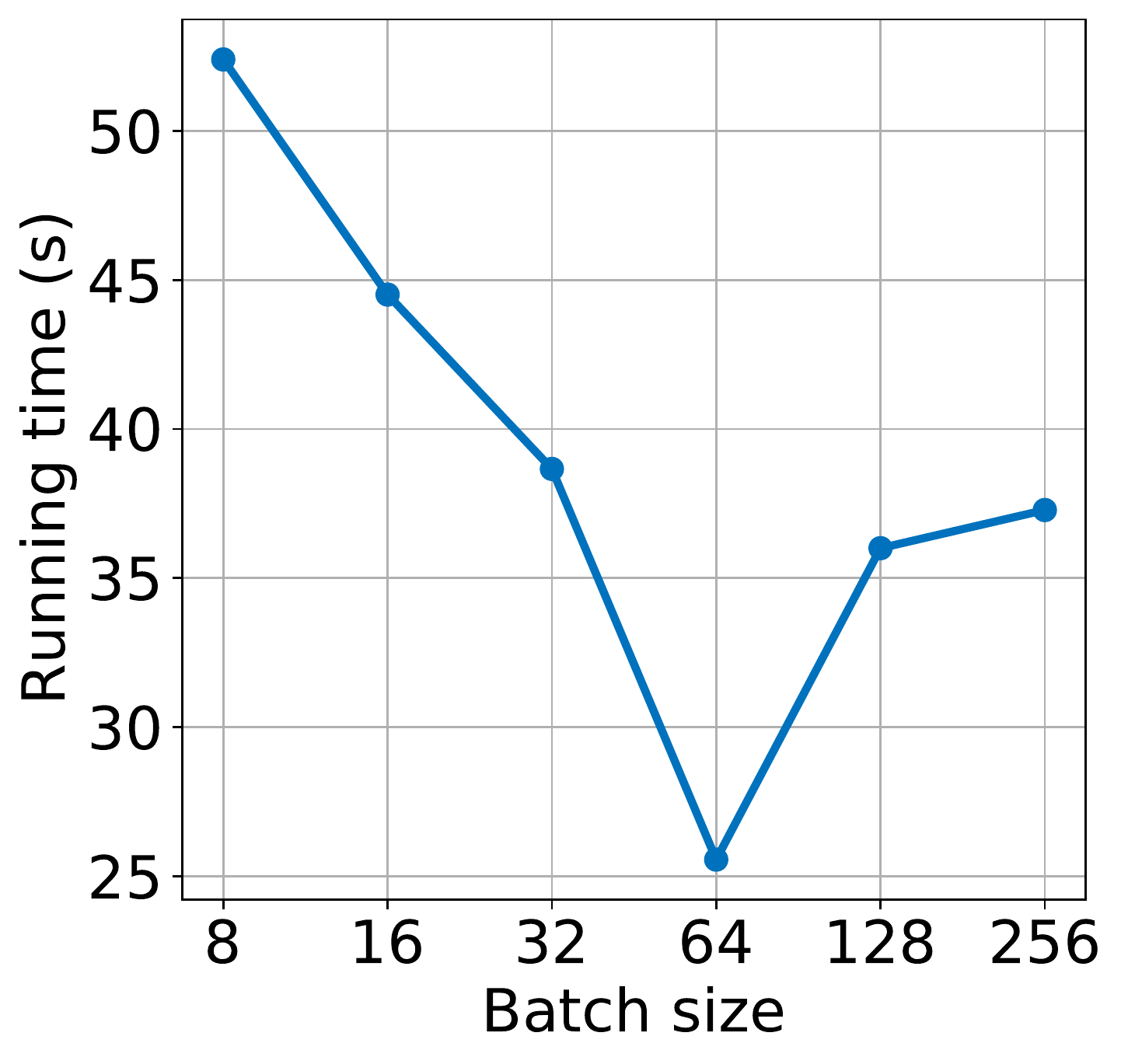}
            \caption{Running time}
            \label{fig:bs_time}
    \end{subfigure}
    \caption{Computational complexity with different batch sizes at test time adaptation. From left to right: accuracy, GPU memory and running time on four domains of PACS \cite{li2017deeper} dataset.}
    \label{fig:efficiency}
    \vspace{-1em}
\end{figure}

\textbf{Efficiency analysis}. Our method adapts a pre-trained model and needs to maintain a memory bank, which may lead to extra computational costs. The usage of GPU memory is mainly influenced by the batch size at test time. We present computational complexity analysis in Fig. \ref{fig:efficiency}. The default batch size of our method is 128 and the reported accuracy is 89.4\% with an almost 14 GB GPU memory cost. While our method still achieves good performance (89.1\% accuracy) when we set the batch size to 64, the GPU memory cost decreases to 7.85 GB and the running time is the fastest for all. When we apply our method to the real world applications, there is a trade-off between accuracy, GPU memory, and running time and we need to balance the choices due to different requirements.

Another strategy for reducing computational complexity is that we only update affine (\ie, scale and shift) parameters when we adapt the model, like \cite{wang2021tent}. We conducted the experiments of updating affine parameters in batch normalization layers, and the result is 89\% accuracy on the PACS dataset. Compared with 89.4\% which is achieved by updating all parameters, the accuracy only decreases by 0.4\% and the GPU memory cost drops by 2 GB. This shows that, with our method, only adapting affine parameters in batch normalization layers can achieve competitive performance with a salience GPU memory decrease.

\subsection{Scalability to Medical Image Segmentation}
Furthermore, we evaluated our algorithm on several medical image segmentation datasets. We conduct the experiments on three tasks: 1) prostate segmentation. We choose Promise12 \cite{litjens2014evaluation} dataset, including 50 cases as source data and the MSD05 dataset (32 cases) from Medical Segmentation Decathlon \cite{Antonelli2022} as test data. 2) cardiac structure segmentation. We choose ACDC \cite{Bernard2018} dataset (200 cases) as source data and LGE images (45 cases) from Multi-sequence Cardiac MR Segmentation Challenge \cite{Zhuang2019,Zhuang2016} as test data. The task is to segment the left ventricular blood pool, right ventricular blood pool, and myocardium. 3) optic disc and cup segmentation. We choose the training set of REFUGE challenge (400 images) \cite{Orlando2020} as the source domain, and RIM-ONE-r3 (159 images) \cite{Fumero2011} and Drishti-GS (101 images) \cite{sivaswamy2015comprehensive} as two different target domains. We aim to segment the optic disc and optic cup. For RIM-ONE-r3 and Drishti-GS, we split them into 99/60 and 51/50 for training and test. For other datasets, we randomly split them to 80\% for training and 20\% for the test.  

\textbf{Implementation.} We adopt DeepLabv3+ \cite{deeplabv3+} with MobileNetV2 \cite{mobilenetv2} backbone as segmentation models. In training and test phases, we use Adam \cite{Adam} optimizer with fixed learning rate as $1e^{-4}$ without specific choice. We enlarge $K$ (\cf. Eq. \ref{eq:spatial_loss}) from 3 to 64 in the segmentation task. We use Dice Score \cite{vnet} as the evaluation metric and we reported the average results of all the target structures in Table \ref{tab:results_medical}. 

\textbf{Results.} We utilize ERM \cite{DBLP:books/daglib/0097035} and Tent \cite{wang2021tent} as the comparison methods. Table \ref{tab:results_medical} shows that our method generally improves the baseline on different tasks. Specifically, we improve ERM baseline 3.9\%, 4.44\%, 3.57\%, and 12.72\% for the four tasks. Experiments prove that our method can also work well in image segmentation tasks. 

\section{Limitations}
Our work has some limitations. First, even though we have already validated our method on several classification and segmentation tasks, we have not extended our method to low-level tasks, such as image dehazing, denoising, and super-resolution tasks. As for low-level tasks, there is no conception of semantic prototype or entropy which can lead to the failure of our method. The second limitation is \textit{hyperparameter search}. In this paper, we do hyperparameter search on the training domain validation set. However, the best choice on the training domains does not mean the best performance on the test domain \cite{rame2022fishr}. We may need to try more strategies to conduct experiments in the TTA setting like test domain validation.

\begin{table}[t]
    \setlength{\belowcaptionskip}{-10pt}
  \centering
  \caption{Results on medical image segmentation. We highlight the \textbf{best} result for each column.}
    \resizebox{\linewidth}{!}
    {
    \begin{tabular}{lcccc}
    \toprule
    Source & MSD05 & ACDC  & \multicolumn{2}{c}{REFUGE} \\
    \midrule
    Target & Promise12 & LGE & DrishtiGS & RIM-ONE-r3 \\
    \midrule
    ERM \cite{DBLP:books/daglib/0097035}   & 83.00  & 82.34 & 81.05 & 68.17 \\
    Tent \cite{wang2021tent}   & 85.64  & 81.11 & 83.96 & 77.87 \\
    Ours  & \textbf{86.90}  & \textbf{86.78} & \textbf{84.62} & \textbf{80.89} \\
    \bottomrule
    \end{tabular}
    }
  \label{tab:results_medical}
  \vspace{-0.5em}
\end{table}
\section{Conclusion}
In this work, we focus on the test time adaptation and propose a new perspective that test time adaptation could be viewed as a feature revision problem. Furthermore, feature revision includes two parts: test time feature uniformity and test time feature alignment. As for the feature uniformity perspective, we propose \textit{Test Time Self-Distillation} to make the target feature as uniform as possible in the adaption. To align features from the same class, we propose \textit{Memorized Spatial Local Clustering} to encourage that the distance between feature representations in the latent space should align with the pseudo logits. A mass of experiments proves that our method not only generally improves the ERM baseline but also outperforms existing TTA or SFDA methods on four domain generalization benchmarks. Furthermore, our method could be adopted to different backbones. To broaden our model for real world applications, we then validate our method on four cross-domain medical image segmentation tasks. Experiment results show that our method is effect, flexible and scalable.

\section*{Acknowledgment}
This work was supported by the National Natural Science Foundation of China, 81971604 and the Grant from the Tsinghua Precision Medicine Foundation, 10001020104.
{\small
\bibliographystyle{ieee_fullname}
\bibliography{egbib}
}
\clearpage
\appendix
This supplementary material provides more details about our implementation (Sec.~\ref{sec:detail}), additional experiments on three corruption datasets (Sec.~\ref{sec:exp_corruption}), some discussion (Sec.~\ref{sec:discussion}), running time analysis (Sec.~\ref{sec:running_time}) and detailed results (Sec.~\ref{sec:full_results}).
\section{Other Implement Details}
\label{sec:detail}
We run our experiments mainly on a single RTX-A6000 or RTX-2080Ti GPU, depending on the need for GPU memory. For source training, we set the batch size as 32 for each source domain and the learning rate as $5e^{-5}$. We set dropout probability and weight decay to zero. We train source model 5k iterations except for DomainNet. We tripled it from 5k to 15k following \cite{swad}. All images are resized to 224$\times$224 and data augmentation is used in source domain training, which includes randomly cropping, flipping horizontally, jittering colour, and changing the intensity.

For implementations of different test time adaptation methods, we use publicly released code of T3A \cite{IwasawaM21}\footnote{\href{https://github.com/matsuolab/T3A}{https://github.com/matsuolab/T3A}}, except for LAME \footnote{\href{https://github.com/fiveai/LAME}{https://github.com/fiveai/LAME}} \cite{BoudiafMAB22} and ETA\footnote{\href{https://github.com/mr-eggplant/EATA}{https://github.com/mr-eggplant/EATA}} \cite{pmlr-v162-niu22a} we use source code of authors. For PL \cite{lee2013pseudo}, we set confidence as 0.9. We resize all images to 224$\times$224 and no data augmentation is used during the test time adaptation process.

For different backbones, we use \texttt{torchvision} implementation \footnote{\href{https://github.com/pytorch/vision}{https://github.com/pytorch/vision}} except for ViT-B/16 and MLP-mixer, we use implementation from \texttt{timm} library\footnote{\href{https://github.com/rwightman/pytorch-image-models}{https://github.com/rwightman/pytorch-image-models}}. For all experiments, we use three random seeds \{0,1,2\} and report the average results in the main text.

\section{Experiments on Corruption Benchmark}
\label{sec:exp_corruption}
We conduct experiments on three corruption datasets~\cite{HendrycksD19}, including CIFAR-10/100-C and ImageNet-C. The results are listed in Table \ref{tab:exp_cifar_imagenet}. From the Table \ref{tab:exp_cifar_imagenet}, it is noticed that our method still achieves the state-of-the-art performance on three corruption datasets.

\section{Discussion}
\label{sec:discussion}
\noindent
\textbf{Performance on VLCS}.  All compared methods shown poor performance on VLCS dataset. We notice that the label distribution is severely different among domains in VLCS compared to other datasets (\eg PACS and OfficeHome), which is probably why fewer methods show performance gain in VLCS compared to other datasets. We calculated the label distribution of VLCS dataset, as shown in Table \ref{tab:label_vlcs}. Since the label distribution is not considered for existing methods, the adaptation may fail.

\noindent
\textbf{Related Work}. There are some papers considering nearest neighbor information in the test time adaptation setting.
TAST \cite{jang2023testtime}, a concurrent work published in ICLR 2023, considers nearest neighbor information to refine pseudo-labels. AdaContrast~\cite{0001WDE22} uses a soft voting strategy among the nearest neighbors in the feature space to refine pseudo-labels. Both of them use nearest neighbor information to refine pseudo-labels. Different from them, our \textit{Memorized Spatial Local Clustering} aims to cluster features with the same pseudo-label.

\section{Running Time Analysis}
\label{sec:running_time}
We provide the running time of different methods for reference. Running time is tested using RTX-A6000 GPU and AMD-EPYC-7542 32 Core Processor. The results are listed in Table \ref{tab:running_time_analysis} using ResNet50 \cite{he2016deep} backbone.

\section{Full Results}
\label{sec:full_results}
We provide full results of different methods on ResNet-18/50, including results for each domain and error bars across three random seeds. See Table \ref{tab:pacs_resnet18}-\ref{tab:domainnet_resnet50}.
\begin{table}[t]
  \centering
  \caption{Label distribution on VLCS dataset.}
    \begin{tabular}{lccccc}
    \toprule
          & 0     & 1     & 2     & 3     & 4 \\
    \midrule
    Caltech101 & 237   & 123   & 118   & 67    & 870 \\
    LabelMe & 80    & 1209  & 88    & 42    & 1237 \\
    SUN09 & 20    & 932   & 1036  & 30    & 1264 \\
    VOC2007 & 330   & 699   & 428   & 420   & 1499 \\
    \bottomrule
    \end{tabular}
  \label{tab:label_vlcs}
\end{table}%

\begin{table}[t]
  \centering
  \caption{Accuracy on CIFAR-10/100-C and ImageNet-C. We use ResNet26 for CIFAR-10/100-C, and ResNet50 for ImageNet-C.}
  \resizebox{\linewidth}{!}{
   \begin{tabular}{lccc}
    \toprule
          & CIFAR-10-C & CIFAR-100-C & ImageNet-C \\
    \midrule
    ERM \cite{DBLP:books/daglib/0097035} & 70.7 & 41.4 & 18.0 \\
    BN \cite{SchneiderRE0BB20} & 77.4 & 48.0 & 33.5 \\
    Tent \cite{wang2021tent} & 80.7 & 51.7 & 42.7 \\
    PL \cite{lee2013pseudo}& 80.2 & 49.8 & 38.4 \\
    SHOT-IM \cite{liang2020we}& 80.7 & 52.1  & 43.1 \\    
    T3A \cite{IwasawaM21}& 77.4 & 44.6 & 36.5 \\     
    ETA \cite{pmlr-v162-niu22a}& 80.6  & \underline{52.4}  & \textbf{48.1} \\
    LAME \cite{BoudiafMAB22}& 79.4  & 50.6  & 47.6 \\
    \hline
    Ours  & \textbf{81.7}  & \textbf{52.6}  & \underline{48.0}\\
    \bottomrule
    \end{tabular}  
  }    
  \label{tab:exp_cifar_imagenet}
\end{table}
\begin{table*}[t]
  \centering
  \caption{Running time analysis of different methods on four datasets. The units used in the table are seconds.  ``GD'' denotes whether the method requires gradient-based optimization.}
  
    \begin{tabular}{lccccc}
    \toprule
          & GD    & PACS \cite{li2017deeper}  & VLCS \cite{torralba2011unbiased}  & OfficeHome \cite{venkateswara2017deep} & DomainNet \cite{PengBXHSW19} \\
    \midrule
    ERM \cite{DBLP:books/daglib/0097035}   &  \XSolidBrush    & 17   & 17   & 56   & 736 \\
    BN \cite{SchneiderRE0BB20}   & \XSolidBrush     & 18   & 18   & 56   & 745 \\
    Tent \cite{wang2021tent}  & \Checkmark  & 32   & 33   & 67   & 1094 \\
    PL \cite{lee2013pseudo}   & \Checkmark     & 33   & 34   & 69   & 1339 \\
    SHOT-IM \cite{liang2020we}& \Checkmark     & 37   & 39   & 72   & 1339 \\
    T3A \cite{IwasawaM21}  & \XSolidBrush     & 19   & 19   & 57   & 829 \\
    ETA \cite{pmlr-v162-niu22a} & \Checkmark     & 21   & 22   & 66   & 1094 \\
    LAME \cite{BoudiafMAB22}  & \XSolidBrush     & 18   & 19   & 56  & 750 \\
    \hline
    Ours   & \Checkmark    & 35   & 37   & 71  & 1644 \\
    \bottomrule
    \end{tabular}
  \label{tab:running_time_analysis}%
\end{table*}

\begin{table*}[t]
  \centering
  \caption{Full results on PACS with ResNet18.}
    \begin{tabular}{lccccc}
    \toprule
    & A   & C & P & S & Avg \\
    \midrule
    ERM & 80.56$\pm$0.45 & 77.36$\pm$0.85 & 93.01$\pm$0.17 & 77.35$\pm$2.90 & 82.07$\pm$0.49 \\
    BN  & 81.60$\pm$0.16 & 82.00$\pm$0.51 & 92.85$\pm$0.24 & 74.86$\pm$1.10 & 82.82$\pm$0.34 \\ 
    Tent& 83.43$\pm$0.53 & 83.02$\pm$0.74 & 93.88$\pm$0.38 & 79.35$\pm$1.15 & 84.92$\pm$0.32\\
    PL  & 84.93$\pm$1.32 & 83.27$\pm$1.78 & 92.62$\pm$1.37 & 77.72$\pm$3.66 & 84.64$\pm$1.13 \\
SHOT-IM & 84.61$\pm$1.06 & 82.36$\pm$1.88 & 93.60$\pm$0.38 & 69.64$\pm$3.40 & 82.55$\pm$1.07 \\
    T3A & 83.00$\pm$0.76 & 79.56$\pm$0.44 & 94.48$\pm$0.34 & 76.95$\pm$2.94 & 83.50$\pm$0.67 \\
    ETA & 81.33$\pm$0.30 & 81.89$\pm$0.55 & 92.82$\pm$0.53 & 74.77$\pm$0.91 & 82.70$\pm$0.31 \\
   LAME & 83.05$\pm$0.53 & 83.06$\pm$0.51 & 94.30$\pm$0.29 & 77.91$\pm$0.80 & 84.58$\pm$0.23 \\
   Ours & 86.50$\pm$0.75 & 86.38$\pm$0.82 & 94.57$\pm$0.32 & 81.84$\pm$0.94 & 87.32$\pm$0.39 \\
    \bottomrule
    \end{tabular}%
  \label{tab:pacs_resnet18}%
\end{table*}%

\begin{table*}[t]
  \centering
  \caption{Full results on OfficeHome with ResNet18.}
    \begin{tabular}{lccccc}
    \toprule
    & A   & C & P & R & Avg \\
    \midrule
   ERM  & 55.49$\pm$0.61 & 51.41$\pm$0.46 & 71.92$\pm$0.47 & 73.67$\pm$0.18 & 63.12$\pm$0.26 \\
    BN  & 54.36$\pm$1.00 & 51.14$\pm$0.25 & 71.20$\pm$0.44 & 72.52$\pm$0.37 & 62.30$\pm$0.25 \\
  Tent  & 55.85$\pm$0.91 & 53.38$\pm$0.29 & 72.50$\pm$0.65 & 73.26$\pm$0.37 & 63.75$\pm$0.23 \\
    PL  & 54.64$\pm$0.96 & 48.36$\pm$1.58 & 68.83$\pm$1.08 & 69.05$\pm$0.58 & 60.22$\pm$0.37 \\
SHOT-IM & 55.45$\pm$1.08 & 52.32$\pm$0.99 & 73.23$\pm$0.76 & 72.67$\pm$0.37 & 63.42$\pm$0.52 \\
    T3A & 56.18$\pm$0.48 & 52.90$\pm$0.67 & 73.44$\pm$0.48 & 74.48$\pm$0.19 & 64.25$\pm$0.22 \\
   ETA  & 54.88$\pm$0.33 & 51.05$\pm$0.21 & 71.18$\pm$0.44 & 72.72$\pm$0.30 & 62.46$\pm$0.14 \\
   LAME & 54.84$\pm$0.86 & 50.90$\pm$0.26 &	70.85$\pm$0.30 & 72.19$\pm$0.30 & 62.20$\pm$0.21 \\ 
   Ours & 57.87$\pm$0.81 & 53.40$\pm$0.27 & 74.20$\pm$0.38 & 73.86$\pm$0.22 & 64.83$\pm$0.46 \\
    \bottomrule
    \end{tabular}%
  \label{tab:officehome_resnet18}%
\end{table*}%

\begin{table*}[t]
  \centering
  \caption{Full results on VLCS with ResNet18.}
    \begin{tabular}{lccccc}
    \toprule
    & C  & L & S & V & Avg \\
    \midrule
    ERM & 92.44$\pm$0.78 & 62.72$\pm$0.69 & 69.26$\pm$2.01 & 66.58$\pm$1.43 & 72.75$\pm$0.29 \\
   BN   & 76.23$\pm$1.99 & 57.84$\pm$0.80 & 58.04$\pm$0.73 & 65.12$\pm$0.35 & 64.31$\pm$0.47 \\
   Tent & 82.65$\pm$1.34 & 60.02$\pm$1.04 & 60.49$\pm$0.84 & 66.28$\pm$0.46 & 67.36$\pm$0.43 \\
    PL  & 86.64$\pm$3.45 & 61.66$\pm$1.47 & 61.78$\pm$2.66 & 65.64$\pm$1.53 & 68.93$\pm$1.07 \\
SHOT-IM & 76.65$\pm$2.94 & 57.34$\pm$1.26 & 59.13$\pm$0.85 & 66.46$\pm$0.55 & 64.90$\pm$0.70 \\
    T3A & 96.76$\pm$1.22 & 63.80$\pm$0.39 & 64.98$\pm$0.45 & 66.55$\pm$0.38 & 73.03$\pm$0.35 \\
    ETA & 76.1$\pm$1.90  & 57.89$\pm$0.87 & 58.12$\pm$0.73 & 65.31$\pm$0.23 & 64.35$\pm$0.40 \\
   LAME & 94.7$\pm$0.21  & 62.69$\pm$0.25 & 67.58$\pm$0.12 & 66.55$\pm$0.45 & 72.88$\pm$0.13 \\
   Ours & 97.2$\pm$1.15  & 64.50$\pm$0.28 & 65.42$\pm$0.34 & 67.32$\pm$0.38 & 73.61$\pm$ 0.42 \\
    \bottomrule
    \end{tabular}%
  \label{tab:vlcs_resnet18}%
\end{table*}%

\begin{table*}[t]
    \centering
    \caption{Full results on DomainNet with ResNet18.}
    \begin{tabular}{lccccccc}
    \toprule
         &  clipart        & infograph      & painting & quickdraw & real & sketch & Avg\\
    \midrule
     ERM & 55.86$\pm$0.15 & 16.85$\pm$0.06 & 44.80$\pm$0.20 & 12.49$\pm$0.38 & 56.74$\pm$0.04 & 46.96$\pm$0.12 & 38.95$\pm$0.08\\
     BN  & 55.90$\pm$0.09 & 12.07$\pm$0.15 & 43.58$\pm$0.06 & 11.63$\pm$0.15 & 56.47$\pm$0.09 & 47.16$\pm$0.15 & 37.80$\pm$0.06 \\
    Tent & 56.67$\pm$0.14 & 13.58$\pm$0.19 & 45.02$\pm$0.13 & 11.52$\pm$0.33 & 57.25$\pm$0.05 & 48.59$\pm$0.16 & 38.77$\pm$0.10\\
     PL  & 55.99$\pm$0.12 & 14.44$\pm$0.27 & 44.29$\pm$0.52 &  4.34$\pm$0.46 & 45.22$\pm$1.31 & 47.09$\pm$0.13 & 35.23$\pm$0.32\\
 SHOT-IM & 56.73$\pm$0.18 & 14.02$\pm$0.22 & 44.61$\pm$0.06 & 16.13$\pm$0.24 & 57.51$\pm$0.13 & 48.20$\pm$0.17 & 39.53$\pm$0.09\\
     T3A & 55.82$\pm$0.18 & 16.71$\pm$0.20 & 43.43$\pm$0.18 & 17.86$\pm$0.26 & 57.58$\pm$0.06 & 46.28$\pm$0.08 & 39.61$\pm$0.04\\
     ETA & 56.46$\pm$0.11 & 14.67$\pm$0.12 & 45.20$\pm$0.10 & 14.13$\pm$0.18 & 57.70$\pm$0.05 & 48.44$\pm$0.14 & 39.43$\pm$0.04\\
    LAME & 55.42$\pm$0.09 & 12.09$\pm$0.16 & 43.35$\pm$0.08 & 11.52$\pm$0.16 & 55.69$\pm$0.09 & 46.86$\pm$0.17 & 37.49$\pm$0.07 \\
    Ours & 56.79$\pm$0.12 & 18.42$\pm$0.14 & 46.71$\pm$0.12 & 13.45$\pm$0.22 & 57.65$\pm$0.12 &	48.12$\pm$0.24 & 40.19$\pm$0.08 \\
    \bottomrule
    \end{tabular}
    \label{tab:domainnet_resnet18}
\end{table*}

\begin{table*}[t]
  \centering
  \caption{Full results on PACS with ResNet50.}
    \begin{tabular}{lccccc}
    \toprule
    & A   & C & P & S & Avg \\
    \midrule
    ERM & 82.50$\pm$1.83 & 80.80$\pm$0.33 & 94.05$\pm$0.30 & 80.99$\pm$1.29 & 84.59$\pm$0.40 \\
    BN  & 83.27$\pm$0.47 & 84.91$\pm$0.43 & 94.03$\pm$0.31 & 77.92$\pm$1.23 & 85.03$\pm$0.20 \\
   Tent & 85.28$\pm$1.07 & 86.75$\pm$0.92 & 94.94$\pm$0.83 & 82.96$\pm$1.20 & 87.48$\pm$0.52 \\
     PL & 83.96$\pm$1.63 & 84.15$\pm$2.91 & 93.82$\pm$1.74 & 78.99$\pm$2.64 & 85.23$\pm$1.70 \\
SHOT-IM & 84.31$\pm$0.63 & 85.74$\pm$0.56 & 94.04$\pm$0.67 & 77.91$\pm$0.94 & 85.50$\pm$0.31 \\
    T3A & 84.07$\pm$0.68 & 82.37$\pm$0.92 & 95.02$\pm$0.27 & 82.72$\pm$1.06 & 86.04$\pm$0.24 \\
    ETA & 83.27$\pm$0.47 & 84.91$\pm$0.43 & 94.03$\pm$0.31 & 77.92$\pm$1.24 & 85.04$\pm$0.20 \\
   LAME & 84.97$\pm$0.77 & 85.50$\pm$0.55 & 95.04$\pm$0.23 & 80.97$\pm$1.09 & 86.62$\pm$0.22 \\
   Ours & 87.68$\pm$0.84 & 88.78$\pm$0.63 & 96.17$\pm$0.37 & 85.01$\pm$1.52 & 89.41$\pm$0.51 \\
    \bottomrule
    \end{tabular}%
  \label{tab:pacs_resnet50}%
\end{table*}%

\begin{table*}[t]
  \centering
  \caption{Full results on OfficeHome with ResNet50.}
    \begin{tabular}{lccccc}
    \toprule
    & A   & C & P & R & Avg \\
    \midrule
    ERM & 60.71$\pm$0.88 & 55.74$\pm$0.79 & 76.18$\pm$0.65 & 76.83$\pm$0.41 & 67.37$\pm$0.06 \\
    BN  & 58.23$\pm$0.76 & 55.62$\pm$0.68 & 75.08$\pm$0.60 & 75.47$\pm$0.35 & 66.10$\pm$0.20 \\
   Tent & 60.55$\pm$0.93 & 58.73$\pm$0.85 & 76.48$\pm$0.53 & 76.07$\pm$0.59 & 67.96$\pm$0.24 \\
     PL & 59.14$\pm$0.93 & 57.29$\pm$0.62 & 76.24$\pm$0.69 & 75.85$\pm$0.28 & 67.13$\pm$0.17 \\
SHOT-IM & 59.20$\pm$0.76 & 57.54$\pm$0.58 & 76.53$\pm$0.44 & 76.27$\pm$0.37 & 67.39$\pm$0.16 \\
    T3A & 61.23$\pm$1.00 & 56.69$\pm$1.11 & 77.95$\pm$0.43 & 77.31$\pm$0.22 & 68.29$\pm$0.21 \\
    ETA & 58.38$\pm$0.80 & 55.78$\pm$0.69 & 75.17$\pm$0.56 & 75.53$\pm$0.34 & 66.21$\pm$0.19 \\
   LAME & 58.67$\pm$0.70 & 55.58$\pm$0.53 & 75.09$\pm$0.65 & 75.40$\pm$0.31 & 66.19$\pm$0.20 \\
   Ours & 62.32$\pm$0.52 & 57.45$\pm$0.71 & 77.48$\pm$0.45 & 77.45$\pm$0.38 & 68.67$\pm$0.14 \\
    \bottomrule
    \end{tabular}%
  \label{tab:officehome_resnet50}%
\end{table*}%

\begin{table*}[t]
  \centering
  \caption{Full results on VLCS with ResNet50.}
    \begin{tabular}{lccccc}
    \toprule
    & C   & L & S & V & Avg \\
    \midrule
    ERM & 94.91$\pm$0.32 & 65.20$\pm$2.35 & 66.52$\pm$1.65 & 69.41$\pm$3.38 & 74.01$\pm$1.32 \\
    BN  & 75.26$\pm$1.15 & 56.85$\pm$0.53 & 60.87$\pm$0.68 & 66.16$\pm$0.59 & 64.78$\pm$0.34 \\
   Tent & 84.75$\pm$2.25 & 60.70$\pm$1.26 & 64.94$\pm$1.40 & 66.42$\pm$0.89 & 69.20$\pm$0.83 \\
     PL & 86.75$\pm$4.25 & 61.19$\pm$2.22 & 63.36$\pm$3.33 & 62.77$\pm$2.84 & 68.52$\pm$1.79 \\
SHOT-IM & 76.54$\pm$1.27 & 55.90$\pm$1.20 & 61.26$\pm$0.71 & 67.24$\pm$0.86 & 65.23$\pm$0.32\\
    T3A & 97.06$\pm$0.30 & 63.96$\pm$0.89 & 67.14$\pm$0.52 & 67.75$\pm$0.31 & 73.98$\pm$0.32 \\
    ETA & 75.27$\pm$1.13 & 56.85$\pm$0.52 & 60.89$\pm$0.67 & 66.16$\pm$0.58 & 64.79$\pm$0.34 \\
   LAME & 96.25$\pm$0.55 & 61.39$\pm$0.40 & 70.25$\pm$0.67 & 67.89$\pm$0.38 & 73.94$\pm$0.14 \\
   Ours & 97.40$\pm$0.41 & 64.89$\pm$0.64 & 68.05$\pm$0.45 & 68.12$\pm$0.43 & 74.52$\pm$0.27\\
    \bottomrule
    \end{tabular}%
  \label{tab:vlcs_resnet50}%
\end{table*}
\begin{table*}[h]
    \centering
    \caption{Full results on DomainNet with ResNet50.}
    \begin{tabular}{lccccccc}
    \toprule
         &  clipart        & infograph      & painting & quickdraw & real & sketch & Avg\\
    \midrule
     ERM  & 64.76$\pm$0.06 & 22.11$\pm$0.11 & 51.77$\pm$0.18 & 13.84$\pm$0.14 & 64.66$\pm$0.21 & 54.04$\pm$0.27 & 45.20$\pm$0.09\\
     BN   & 64.46$\pm$0.09 & 15.62$\pm$0.05 & 50.64$\pm$0.08 & 11.84$\pm$0.05 & 63.86$\pm$0.11 & 53.86$\pm$0.19 & 43.38$\pm$0.03\\
     Tent & 65.78$\pm$0.08 & 18.18$\pm$0.02 & 52.96$\pm$0.01 & 10.77$\pm$0.11 & 64.85$\pm$0.09 & 55.71$\pm$0.09 & 44.71$\pm$0.03\\
     PL   & 64.96$\pm$0.05 & 19.00$\pm$0.03 & 50.30$\pm$0.27 &  4.21$\pm$0.68 & 54.40$\pm$0.55 & 54.20$\pm$0.15 & 41.18$\pm$0.13\\
SHOT-IM   & 65.62$\pm$0.05 & 18.73$\pm$0.21 & 52.41$\pm$0.08 & 19.01$\pm$0.24 & 66.47$\pm$0.12 & 55.54$\pm$0.08 & 46.30$\pm$0.07\\
    T3A   & 64.77$\pm$0.05 & 22.10$\pm$0.09 & 50.89$\pm$0.15 & 19.41$\pm$0.18 & 65.85$\pm$0.06 & 53.96$\pm$0.17 & 46.16$\pm$0.03\\
    ETA   & 65.11$\pm$0.09 & 19.37$\pm$0.19 & 52.69$\pm$0.11 & 18.24$\pm$0.33 & 65.92$\pm$0.10 & 55.48$\pm$0.16 & 46.13$\pm$0.08\\
    LAME  & 64.18$\pm$0.12 & 15.64$\pm$0.07 & 50.54$\pm$0.04 & 11.77$\pm$0.05 & 63.46$\pm$0.08 & 53.65$\pm$0.18 & 43.20$\pm$0.03\\
    Ours  & 66.12$\pm$0.08 & 24.12$\pm$0.12 & 52.82$\pm$0.10 & 18.17$\pm$0.08 &	68.45$\pm$0.12 & 56.72$\pm$0.12 & 47.73$\pm$0.05\\
    \bottomrule
    \end{tabular}
    \label{tab:domainnet_resnet50}
\end{table*}
\end{document}